\def\year{2019}\relax
\documentclass[letterpaper]{article} 
\usepackage{aaai19}  
\usepackage{times}  
\usepackage{helvet} 
\usepackage{courier}  
\usepackage[hyphens]{url}  
\usepackage{graphicx} 
\usepackage{latexsym}
\usepackage{booktabs}
\usepackage{url}

\usepackage{graphicx}  
\nocopyright
\title{A Research Platform for Multi-Robot Dialogue with Humans}
\author{\Large \textbf{Matthew Marge, Stephen Nogar, Cory J. Hayes, Stephanie M. Lukin,} \\ \Large \textbf{Jesse Bloecker, Eric Holder, Clare Voss}\\ 
U.S. Army Research Laboratory\\
Adelphi, MD 20783\\
matthew.r.marge.civ@mail.mil 
}

\begin{document}

\maketitle

\begin{abstract}
This paper presents a research platform that supports spoken dialogue interaction
with multiple robots. The demonstration showcases our crafted MultiBot testing scenario in which users can
verbally issue search, navigate, and follow instructions to two robotic teammates:
a simulated ground robot and an aerial
robot. This flexible language and robotic platform takes advantage of existing tools for speech recognition and dialogue management that are compatible with new domains, and implements an inter-agent communication protocol (tactical behavior specification), where verbal instructions are encoded for tasks assigned to the appropriate robot.
\end{abstract}

\section{Introduction}
We investigate dialogue dynamics between a human and multiple robotic teammates in scenarios
that reflect the urgency of tasks, such as search and
rescue. While multi-participant dialogue has long been
studied~\cite{sacks1978simplest}, human-robot communication remains an
area that merits further investigation. Natural language offers human teammates
a hands-free way to interact with multiple robots, unlike \textit{direct teleoperation} where a robot is
controlled with a hand-held device.
Language also encourages human teammates to issue complex, abstract-level instructions, rather than lower-level command-and-control instructions.

This demonstration shows the completed integration of spoken dialogue with
simulated robots in a simulated environment. A ground robot (Clearpath Husky) and a small, quadrotor aerial robot (Qualcomm Snapdragon Flight) work with a human teammate in a search and rescue  scenario using a platform we call MultiBot. Spoken instructions are interpreted and independent tasks are delegated by a dialogue manager to these robots,
including waypoint navigation, exploration, and object detection,
as well as joint tasks such as following one another. The
robots adapt their behavior to meet human-specified conditions,
e.g., moving quickly given an urgent task.

One contribution of this research is a software platform for exploring
new aspects of multi-robot dialogue to explore language, robotics,
and human-robot interaction research.
Natural language serves as a single interface to
consistently interact with multiple robots on a team, despite these robots
having different capabilities.
The robotic behaviors are built on
top of the Robot Operating System (ROS), an open-source framework
that uses the same communication protocols for physical platforms
as in simulation, to explore how interaction dynamics change in new environments
and tasks.

A second contribution of this research is methodological:
the software components within our platform
for dialogue management and robot navigation are ``wizard-swappable'',
meaning that human operators (i.e., wizards) may stand-in functionally for
software components yet to be developed, typically in the initial
design stage of an autonomous system.
This approach supports collecting training data for the underlying
algorithms.
The most immediate application of our research platform is
for collecting natural language data from experiment
participants to use in modeling how humans communicate with
multiple robots in task-oriented dialogue.

\section{Scenarios and Research Approach}
\label{approach}

\begin{table*}[t]
\centering
\small
\begin{tabular}{p{0.58in} ||p{0.7in}|p{0.7in} |p{0.7in} |p{0.75in} || p{0.75in} | p{0.7in} }
\hline
& {\bf Exp 1} 		& {\bf Exp 2} 		& {\bf Exp 3}		& {\bf Exp 4}  		&	{\bf ScoutBot}		& {\bf MultiBot	} \\
& \cite{marge2016assessing}  		&  \cite{bonial2017laying}  	& completed 2018  		&  ongoing data collection			&	 \cite{lukin2018scoutbot} 		&  current  \\ \hline \hline

Dialogue  	& wizard + & wizard +  & wizard + 	& ASR + &  ASR + & ASR +\\
Processing & typing & button presses &  button presses & auto-DM & auto-DM & auto-DM  \\ \hline

Robotic  		& wizard +	& wizard +	&  wizard	+ &  wizard +	& finite state	& auto-assign  	\\

Behaviors	& joystick		& joystick		&  joystick	&  joystick		& machine 	&  via TBS 	\\ \hline

Robot(s)  	& 1 physical	& 1 physical	&  1 simulated 	& 1 simulated   		& 1 simulated		& 2 simulated	\\ \hline

Environment & indoors +  	& indoors +	& indoors + 	& indoors +		& indoors +	& outdoors + \\
			& real building	& real building	& sim building	& sim building		& sim building	& sim buildings	 \\	\hline

\end{tabular}
\caption{Testing scenarios over time. Columns depict progression of testing scenario experimentation and development; rows represent scenario components (DM: Dialogue Management; TBS: tactical behavior specification)}
\label{progression}
\end{table*}

Our platform allows for the testing and implementation of new testing scenarios for language, robotics, and human-robot interaction research. A testing scenario may vary the number and types of robots and robotic capabilities, language understanding and dialogue management capabilities, as well as the physical or simulated environment and user task. Scenarios can be crafted for the following interdisciplinary research objectives:\\
\noindent {\it Language research}, to test linguistically-challenging instructions that are situationally-relevant given different objects, structures, landmarks, and locations within the environment;

\noindent {\it Robotics research}, to demonstrate robotic behaviors on the ground and in the air, feasible for the physical robots and their simulated counterparts;

\noindent {\it Human-robot interaction research}, to track effective navigation and coordinated exploration in the environment by multiple autonomous robots as a result of spoken dialogue with human teammates.

The MultiBot Platform was built upon a foundation of prior research and development efforts in these areas of research. The columns of Table~\ref{progression} show the temporal progression and development of various testing scenarios leading up to MultiBot. Technical details specific to MultiBot follow in Section~\ref{details}.

Components supported by the MultiBot Platform are depicted as rows in Table~\ref{progression}. The key milestones in the progression included using human wizards through Experiments 1-4 (prior and ongoing work) with different methods of performing the task (typing or pressing buttons that have predefined text messages) to build up the databases of dialogue interactions, a dialogue manager, and robot behaviors. The data collected in these prior experiments was used to train the ScoutBot system as an end-to-end, fully autonomous dialogue management and autonomous robot implementation \cite{lukin2018scoutbot} for control of a single simulated robot in an indoor simulated building.

The crafting of a new, realistic testing scenario for dialogue with more than one autonomous robot made evident new integration requirements in our research platform. We needed a testing scenario with a coherent, structured narrative involving a ground and an aerial robot demonstrating new behavior sequences as the software components were being developed.
To develop MultiBot, we used a simulated outdoor environment that covers a complex region of roads and buildings.  MultiBot also builds on the ScoutBot testing scenario with two other significant changes: (i)~the human operator can now address these two robots,
each of which may verbally respond with their own feedback, and
(ii)~the navigation commands to the robot from the dialogue manager are now adaptable, pivoting through an intermediary computational representation language that can in turn be mapped to robot-specific behaviors.

The MultiBot Platform supports targeted evaluation of the individual components (the rows in Table~\ref{progression}) as well as holistic, systematic evaluation.
Possible measures of these components include evaluation of the robot behaviors (e.g., distance to goal), the performance of the robot itself (e.g., energy efficiency in performing tasks), and dialogue processing (e.g., coverage of utterances and appropriateness of recovery strategies).
Holistic evaluation in human subject experiments
can measure perceived task workload, success, and satisfaction, as well as overall task completion and efficiency both for individual robots and as a team in comparison to a system using direct teleoperation of the robots.

\section{Architecture Overview}
\label{details}

\begin{figure}[t]
	\centering
  	\includegraphics[width=3.0in]{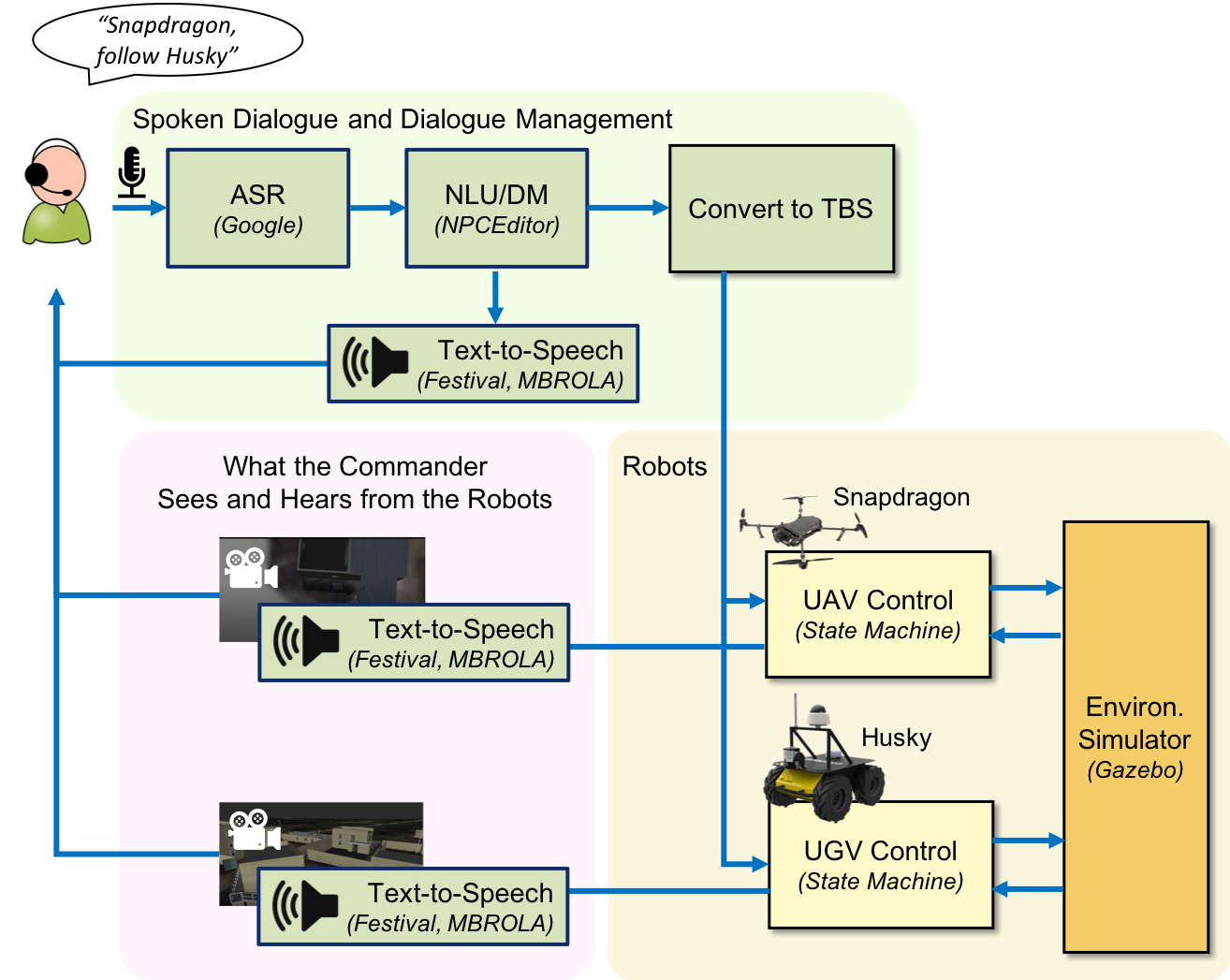}
  	\caption{Platform Architecture, MultiBot Testing Scenario}
  	\label{fig:arch}
\end{figure}

Figure~\ref{fig:arch} showcases a high-level view of the flow of information in our platform, depicting the MultiBot testing scenario as an example. The user visually observes the simulated environment, listens to audio feedback from the robots (lower left-hand corner of Figure~\ref{fig:arch}), and speaks verbal instructions.
A speech recognizer (top left-hand corner of Figure~\ref{fig:arch}) passes the user utterance to a Natural Language Understanding (NLU) and Dialogue Management (DM) process.
The NLU/DM module ensures the utterance is well-formed and executable, and may prompt for additional information from the user. Well-formed instructions are then sent to a process that formats them into an unambiguous, structured command that triggers robot behaviors (lower right-hand corner) which work in tandem with the environment simulator to continuously update the status of the robot(s) within the simulated environment. The platform generates verbal feedback from the robots, and the visualization is updated based on the actions of the robot(s).

\subsection{Spoken Dialogue and Dialogue Management}
The MultiBot Platform is designed to support experimentation through the swapping of different language understanding and processing capabilities. The capabilities chosen for the MultiBot testing scenario leverage existing components of a spoken dialogue and dialogue management interface, extending these tools to support conversing with multiple robots. Speech recognition is supported by Google's Automatic
Speech Recognition API. Dialogue processing utilizes the NPCEditor dialogue
manager~\cite{LeuskiTraum2011,hartholt2013}.
The NLU/DM module interprets dialogue instructions and
produces responses using statistical retrieval algorithms from prior dialogue system implementations \cite{traum2015new,lukin2018scoutbot} which allow for a range of unconstrained speech input.
This testing scenario uses a novel configuration with five categories
of instructions: (1)
wake (get a particular robot's attention), (2) waypoint navigation of
one or more robots, (3) follow-behind commands, (4) inspection,
and (5) patrol of a pre-defined area.

We use the simultaneous message
generation that the ScoutBot system implements to support  generating clarification responses to the human teammate and to the robotic platforms, the latter of which are converted into an instruction issued using a Tactical Behavior Specification (TBS) message \cite{Oh-2015-5903} for MultiBot. This provides a common format for issuing a high level action with relevant location data and object information.
When robot actions are completed, a text message signal is sent that may either be converted from text to speech, as depicted in Figure~\ref{fig:arch}, or shown in a chat window if the environment is noisy or if stealth is desired. Text-to-speech synthesis of the robots and the NLU/DM module is performed using the Festival Speech Synthesis System\footnote{\url{http://www.cstr.ed.ac.uk/projects/festival/}} with MBROLA voices.\footnote{\url{http://tcts.fpms.ac.be/synthesis/mbrola.html}}

\subsection{Robotic Behaviors}
Robotic behaviors tailored for a particular task may be substituted using the MultiBot Platform. For the teaming application supported by the MultiBot testing scenario, robots need the ability to make independent decisions once a verbal instruction has been issued. To support complex actions, we implement a behavior tree based on the open-source Smach library\footnote{\url{http://wiki.ros.org/smach}} within ROS.\footnote{\url{http://www.ros.org/}} The library is an implementation of a finite state machine that manages the robot's behavior, chaining simple actions into more complex actions or tasks. Each robot state (e.g., searching, following, landing) must terminate with one of multiple specific outcomes (e.g., succeeded, failed, interrupted). This outcome determines the next action according to the behavior tree. Once an instruction is implemented as a chain of actions, it can be used as a building block in other instruction, providing a framework for more advanced behaviors.

\begin{figure}[t]
	\centering
  	\includegraphics[width=2.0in]{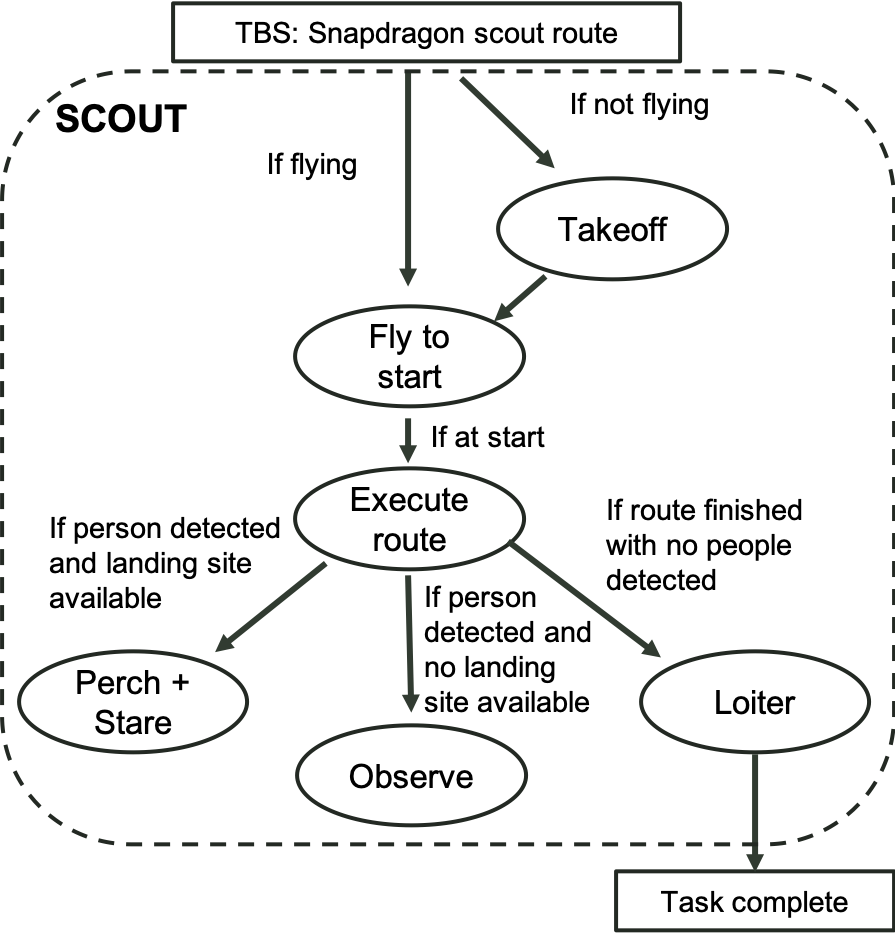}
  	\caption{Behavior tree for ``scout'' instruction}
  	\label{fig:state-machine}
\end{figure}

TBS messages are defined in ROS in coordination with the Smach-based behavior trees. The output of the behaviors are any discovered objects of interest and the resulting state of the robot, such as position, and are provided back to the human operator through the speech synthesis previously described. Included behaviors are ``go-to'', ``follow'', ``scout'', ``search'', and (for the aerial robot) ``takeoff/land''.  The ``scout'' instruction is presented in Figure~\ref{fig:state-machine}. As the robot moves, it uses an onboard camera and an object classifier \cite{bjelonicYolo2018} to search for objects of interest. In the MultiBot testing scenario, if it recognizes an injured person it can consider multiple actions on how best to continue observation. If a suitable landing location is nearby, the robot will execute a ``perch and stare'' behavior, otherwise it will hover nearby and observe.

\section{Demo Summary}
\label{demo}

A demonstration\footnote{A video recording of the testing scenario can be found at \mbox{{\url{https://youtu.be/5kVvj9xEK3E}}}} of the developed behaviors for the MultiBot testing scenario was performed in a software in the loop simulation (SITL). This simulates the robotic sensors and actuators, allowing for verification of the developed language comprehension, perception algorithms, and behavior trees by maximizing the similarity of the code between the robot and the simulator. The demonstration showcases the natural language interface and robot capabilities in a search scenario in which the ground and aerial robotic teammates, named Husky and Snapdragon respectively, are given verbal instructions from a Commander to explore the environment and identify injured individuals.

In the live demonstration, visitors will engage with the two simulated robots in a game scenario to coordinate and navigate both robots to a designated zone along a route with injured individuals. Visitors will be given a 2D map of the environment as well as a set of navigation functions and robot names. The robots will follow the visitor's instructions autonomously, allowing the visitor to analyze the entire map and plan the best route for the robots while simultaneously overcoming several challenges in the task and environment.

\section{Ongoing Research}
\label{ongoing}

This research platform has been used to investigate multi-participant dialogue,
with a particular focus on interaction between one human and multiple robots.
The testing scenario presented in this demo utilizes an \textit{explicit addressee} approach, where tasks
require first
getting an individual robot's attention before issuing a task.
In our ongoing research, we explore an \textit{implicit addressee} approach by leveraging
the NPCEditor dialogue retrieval algorithm, which matches responses to
instructions using a word co-occurrence metric. This required creating
a synthetic dataset of possible instructions. By associating tasks
to robots directly in the training data, the dialogue retrieval
algorithm can automatically match tasks specific to one robot without
requiring mention of that robot's name. For example, tasks related to
flying, such as \textit{``Scout route bravo''}, will be automatically bound
to the aerial robot when the instruction is passed as a TBS. No commonsense
reasoning is required for this capability.
In this research, robot capabilities have been
shown to disambiguate which robot performs which task implicitly,
an improvement over the explicit addressee approach and allows further research on multi-participant dialogue.

Another application of this platform is to categorize user preferences in instructed robot navigation behavior to reduce user workload and improve task efficiency \cite{hayesrss2018}.
The aim is to reflect individual user preferences by automatically fine-tuning robot movements as the interaction history between the user and robot grows, thereby reducing the need for users to verbally provide instruction clarifications or corrections.
The first stage of this research implements inverse reinforcement learning to train a general, automated navigation model from manual human demonstrations.
The second stage will use speech as part of a reward signal to modify the general navigation model on a per-user basis using traditional reinforcement learning techniques.

\section{Related Work}
Several other architectures have explored dialogue with robots. TeamTalk~\cite{marge2019miscommunication} for example,
controls multiple ground robots by way of a predefined grammar, while DIARC~\cite{scheutz2019overview} also supports dialogue with multiple robots, and has been implemented on ground, aerial, and social robots.
Open-source architectures such as OpenDial~\cite{lison2016opendial}, IrisTK~\cite{skantze2012iristk}, and
Microsoft's PSI~\cite{bohus2017rapid} can be used to build many situated
dialogue agents, including robots.
Compared to similar architectures, MultiBot leverages
wizard-swappable components from ScoutBot and extends the mode
of interaction
to multi-participant dialogue. Training data for a
MultiBot-based system can be collected from Wizard-of-Oz
studies or synthetically derived and easily incorporated into MultiBot to accommodate a new
domain, feature, or capability, as Table~\ref{progression} details.

\section{Summary and Future Work}
This paper presents a platform to conduct spoken dialogue interaction with robots in a flexible, scenario-based architecture.
The demonstrated testing scenario, MultiBot, is an implementation of autonomous dialogue management and navigation of two simulated robots in a large, outdoor simulated environment.
This platform enables the crafting of various testing scenarios to perform language, robotics, and human-robot interaction research in a physical or simulated environment with multiple robots, while testing various language and robotic behavior capabilities. The platform provides the opportunity to study human-robot communication and behaviors in competitive and cooperative teaming and train future human-robot teams for a variety of challenging environments.
Additionally, the platform may be used to experiment with new tasks in simulation as new autonomous robot navigation capabilities emerge.

\section*{Acknowledgments}
This work was supported by the U.S. Army Research Laboratory. The authors thank the anonymous reviewers for their feedback, as well as Judith Klavans, Chris Kroninger, and Garrett Warnell for their contributions to this project.

\bibliography{ai-hri2019demo-dsi}

\begin{thebibliography}{}

\bibitem[\protect\citeauthoryear{Bjelonic}{2018}]{bjelonicYolo2018}
Bjelonic, M.
\newblock 2018.
\newblock {YOLO ROS}: {Real-Time Object Detection for} {ROS}.
\newblock \url{https://github.com/leggedrobotics/darknet_ros}.

\bibitem[\protect\citeauthoryear{Bohus, Andrist, and
  Jalobeanu}{2017}]{bohus2017rapid}
Bohus, D.; Andrist, S.; and Jalobeanu, M.
\newblock 2017.
\newblock {Rapid Development of Multimodal Interactive Systems: A Demonstration
  of Platform for Situated Intelligence}.
\newblock In {\em Proceedings of the 19th ACM International Conference on
  Multimodal Interaction}.

\bibitem[\protect\citeauthoryear{Bonial \bgroup et al\mbox.\egroup
  }{2017}]{bonial2017laying}
Bonial, C.; Marge, M.; Artstein, R.; Foots, A.; Gervits, F.; Hayes, C.~J.;
  Henry, C.; Hill, S.~G.; Leuski, A.; Lukin, S.~M.; Moolchandani, P.; Pollard,
  K.~A.; Traum, D.; and Voss, C.~R.
\newblock 2017.
\newblock {Laying Down the Yellow Brick Road: Development of a Wizard-of-Oz
  Interface for Collecting Human-Robot Dialogue}.
\newblock In {\em Proceedings of the AAAI Fall Symposium Series: Natural
  Communication for Human-Robot Collaboration}.

\bibitem[\protect\citeauthoryear{Hartholt \bgroup et al\mbox.\egroup
  }{2013}]{hartholt2013}
Hartholt, A.; Traum, D.; Marsella, S.~C.; Shapiro, A.; Stratou, G.; Leuski, A.;
  Morency, L.-P.; and Gratch, J.
\newblock 2013.
\newblock {All Together Now: Introducing the Virtual Human Toolkit}.
\newblock In {\em Proceedings of the International Conference on Intelligent
  Virtual Agents}.

\bibitem[\protect\citeauthoryear{Hayes \bgroup et al\mbox.\egroup
  }{2018}]{hayesrss2018}
Hayes, C.~J.; Marge, M.; Stump, E.; Bonial, C.; Voss, C.; and Hill, S.~G.
\newblock 2018.
\newblock {Towards Learning User Preferences for Remote Robot Navigation}.
\newblock In {\em Proceedings of the RSS 2018 Workshop on Models and
  Representations for Human-Robot Communication}.

\bibitem[\protect\citeauthoryear{Leuski and Traum}{2011}]{LeuskiTraum2011}
Leuski, A., and Traum, D.
\newblock 2011.
\newblock {NPCEditor: Creating Virtual Human Dialogue Using Information
  Retrieval Techniques}.
\newblock {\em AI Magazine} 32(2).

\bibitem[\protect\citeauthoryear{Lison and
  Kennington}{2016}]{lison2016opendial}
Lison, P., and Kennington, C.
\newblock 2016.
\newblock {OpenDial: A Toolkit for Developing Spoken Dialogue Systems with
  Probabilistic Rules}.
\newblock In {\em Proceedings of the Annual Meeting of the Association for
  Computational Linguistics - System Demonstrations}.

\bibitem[\protect\citeauthoryear{Lukin \bgroup et al\mbox.\egroup
  }{2018}]{lukin2018scoutbot}
Lukin, S.~M.; Gervits, F.; Hayes, C.~J.; Leuski, A.; Moolchandani, P.; Rogers,
  J.~G.; Amaro, C.~S.; Marge, M.; Voss, C.; and Traum, D.
\newblock 2018.
\newblock {ScoutBot: A Dialogue System for Collaborative Navigation}.
\newblock In {\em Proceedings of the Annual Meeting of the Association for
  Computational Linguistics - System Demonstrations}.

\bibitem[\protect\citeauthoryear{Marge and
  Rudnicky}{2019}]{marge2019miscommunication}
Marge, M., and Rudnicky, A.~I.
\newblock 2019.
\newblock {Miscommunication Detection and Recovery in Situated Human-Robot
  Dialogue}.
\newblock {\em ACM Transactions on Interactive Intelligent Systems} 9(1).

\bibitem[\protect\citeauthoryear{Marge \bgroup et al\mbox.\egroup
  }{2016}]{marge2016assessing}
Marge, M.; Bonial, C.; Pollard, K.~A.; Artstein, R.; Byrne, B.; Hill, S.~G.;
  Voss, C.; and Traum, D.
\newblock 2016.
\newblock {Assessing Agreement in Human-Robot Dialogue Strategies: A Tale of
  Two Wizards}.
\newblock In {\em Proceedings of the International Conference on Intelligent
  Virtual Agents}.

\bibitem[\protect\citeauthoryear{Oh \bgroup et al\mbox.\egroup
  }{2015}]{Oh-2015-5903}
Oh, J.; Suppe, A.; Duvallet, F.; Boularias, A.; Vinokurov, J.; Navarro-Serment,
  L.~E.; Romero, O.; Dean, R.; Lebiere, C.; Hebert, M.; and Stentz, A.~T.
\newblock 2015.
\newblock {Toward Mobile Robots Reasoning Like Humans}.
\newblock In {\em Proceedings of the Twenty-Ninth AAAI Conference on Artificial
  Intelligence}.

\bibitem[\protect\citeauthoryear{Sacks, Schegloff, and
  Jefferson}{1978}]{sacks1978simplest}
Sacks, H.; Schegloff, E.~A.; and Jefferson, G.
\newblock 1978.
\newblock {A Simplest Systematics for the Organization of Turn-Taking for
  Conversation}.
\newblock {\em Langauge} 50(4).

\bibitem[\protect\citeauthoryear{Scheutz \bgroup et al\mbox.\egroup
  }{2019}]{scheutz2019overview}
Scheutz, M.; Williams, T.; Krause, E.; Oosterveld, B.; Sarathy, V.; and Frasca,
  T.
\newblock 2019.
\newblock {An Overview of the Distributed Integrated Cognition Affect and
  Reflection DIARC Architecture}.
\newblock In {\em Cognitive Architectures}.

\bibitem[\protect\citeauthoryear{Skantze and
  Al~Moubayed}{2012}]{skantze2012iristk}
Skantze, G., and Al~Moubayed, S.
\newblock 2012.
\newblock {IrisTK: A Statechart-Based Toolkit for Multi-Party Face-to-Face
  Interaction}.
\newblock In {\em Proceedings of the 14th ACM International Conference on
  Multimodal Interaction}.

\bibitem[\protect\citeauthoryear{Traum \bgroup et al\mbox.\egroup
  }{2015}]{traum2015new}
Traum, D.; Jones, A.; Hays, K.; Maio, H.; Alexander, O.; Artstein, R.; Debevec,
  P.; Gainer, A.; Georgila, K.; Haase, K.; Jungblut, K.; Leuski, A.; Smith, S.;
  and Swartout, W.
\newblock 2015.
\newblock {New Dimensions in Testimony: Digitally Preserving a Holocaust
  Survivor's Interactive Storytelling}.
\newblock In {\em International Conference on Interactive Digital
  Storytelling}.

\end{thebibliography}
\bibliographystyle{aaai}

\end{document}